\pdfoutput=1 % ensure pdflatex (for arXiv)

\documentclass[11pt,a4paper]{article}
\usepackage[hyperref]{acl2020}
\usepackage{nert}
\usepackage{times}
\usepackage{dialogue}
\usepackage{latexsym}
\usepackage{microtype}

\aclfinalcopy % Uncomment this line for the final submission
\title{(Re)construing Meaning in NLP}

\author{Sean Trott \\
  University of California, San Diego \\
   \eml{sttrott@ucsd.edu} 
  \And
Tiago Timponi Torrent\\
  Federal University of Juiz de Fora  \\
  \eml{tiago.torrent@ufjf.edu.br} \\
  \AND
Nancy Chang\\
  Google \\
  \eml{ncchang@google.com} \\
  \And
Nathan Schneider\\
  Georgetown University \\
  \eml{nathan.schneider@georgetown.edu} \\
}
\date{}

\begin{document}
\maketitle
\begin{abstract}
Human speakers have an extensive toolkit of ways to express themselves. In this paper, we engage with an idea largely absent from discussions of meaning in natural language understanding—namely, that the way something is expressed reflects different ways of conceptualizing or \emph{construing} the information being conveyed. We first define this phenomenon more precisely, drawing on considerable prior work in theoretical cognitive semantics and psycholinguistics. We then survey some dimensions of construed meaning and show how insights from construal could inform theoretical and practical work in NLP. %language use and practical applications. language resources and technologies. \st{Revised based on TT's suggestions (see below).}

%We then survey some mechanisms of construal that we believe are worthy of attention in NLP for building practical language resources and technologies, as well as for understanding the distinctions between human and artificial linguistic intelligence.

%\ttor{Needs revision upon completion of NC's final final final final pass}\st{Was there something specific you had in mind? I think it looks pretty decent.}\ttor{I think we don't survey the mechanisms of construal, but its dimensions. Also, we provide examples of phenomena in which different dimensions are in action. Finally, and only then, based on 3 case studies, we discuss why construal is worthy of attention in NLP} 
\end{abstract}

\section{Introduction}

Natural language is a versatile tool for allowing humans to express all manner of communicative intents, from simple descriptions of the entities and situations in their direct experience to elaborate rhetorical flights of fancy. Many NLP applications, such as information extraction, question answering, summarization, and dialogue systems, have restricted their scope to what one might call objective information content---relatively uncontroversial facts that systems can infer from an utterance, store in a database and reason about.

While it is tempting to equate such information with the meaning of an utterance, a large body of literature in linguistics and psycholinguistics argues that an utterance conveys much more than a simple set of facts: it carries with it a halo of intimations arising from the speaker's choices, including considerations of perspective, emphasis, and framing. That is, \emph{linguistic choices subtly color meaning}; far from merely conveying objective facts, they reflect how speakers conceptualize meaning and affect listeners' interpretations in predictable ways.

%\nc{Shorten below / replace with catchier opening example.}\nss{``Nora runs her house with an iron fist''? house = metonymy, subject of runs = she is in control, iron fist = tyrant metaphor. Explain that Nora is an infant who has been variously construed as a hard-won prize, a hero, a tyrant, a coworker ...}\st{I like it! Maybe we could have an oppositional/alternate version too (e.g., construing Nora as a hero/hard-won prize, to illustrate how the same entity/set of events can be construed multiple ways?}

Take, for example, this metaphor-rich portrayal of a newborn as a tyrant over her parental subjects:
%\nss{Good, this gives an example imagery and storytelling that should be relevant to meaning as opposed to simple dry facts}
%\ex.\label{ex:nora}
%Life under Nora’s adorable tyranny was filled with squawking, swaddling and ceaseless sleep-input-output cycles until she relaxed her tiny iron grip.

\ex.\label{ex:nora}
Nora's arrival brought a regime change.
Life under her adorable tyranny was filled with squawking, swaddling and ceaseless sleep-input-output cycles. We were relieved when she relaxed her tiny iron grip.
    
This report of new parenthood describes a major life change along with everyday caregiver routines, but its emphasis is on the parents' experience of being suppressed (\textit{under}) and controlled (\textit{grip}) by a creature who is cast, variously, as a tyrant (\textit{regime}), a bird (\textit{squawk}), and a relentless machine (\textit{sleep-input-output cycles}, \textit{iron grip})---albeit a (subjectively) adorable one. 

%\nss{this illustrates metaphor---any other types of construal?}\nc{Decided against mentioning specific technical terms, since it'll be easy to refer back to these in the sections below instead. Is this enough?}

%\nc{Should we include a shortened version of original examples? They support the variability point.}\nss{Yeah. I think my favorites are b, d, and e.}

The power of linguistic choices to shape understanding is also evident in more mundane (and well-studied) examples:
\ex.\label{ex:intro}
    %\a.\label{ex:intro1} The painting is \textit{above} the piano. \\The piano is \textit{below} the painting.
    \a.\label{ex:intro2} Chuck \textit{bought} a car \textit{from} Jerry.\\ Jerry \textit{sold} a car \textit{to} Chuck.\\ Jerry \textit{paid} Chuck \textit{for} the car.
    %\b.\label{ex:msft-voice} Microsoft \textit{employs} me.\\ I \textit{am employed by} Microsoft.
    \b.\label{ex:msft-prep} I work \textit{at} Microsoft.\\ I work \textit{for} Microsoft.
    \b.\label{ex:intro5} The statue \textit{stands} in the plaza. \\The statue \textit{is standing} in the plaza.
    %\b.\label{ex:intro6} The road \textit{runs} through the forest.\\ The road \textit{meanders} through the forest.
    \z.

%\nc{to update for example. But note that current text (and next paragraph) relies on the idea of variation, which the new example no longer has.}\st{Hmm, good point. I do like the idea of emphasizing variable angles on the same scene. Maybe we can emphasize/add various construals to the Nora example? The current example at least already construes her alternately as a tyrant, bird, and machine, so maybe that's enough---what I like about that is metaphor is quite clear, and I think most people will get it, and then the rest of the paper dives into even more subtle distinctions.}\nc{Keep and shorten examples, in progress.}

Each set includes sentences that convey roughly the same facts---i.e.~they could describe the same scenario---but nonetheless differ in various respects. The familiar \textit{framing} differences between buy\slash sell\slash pay \cref{ex:intro2} focus attention on different participants and subevents in a commercial transaction. 
\cref{ex:msft-prep} involves a subtler difference in emphasis, where the choice of \textit{at} highlights the location of the work, while \textit{for} evokes how that work benefits the employer. Grammatical marking can also shift event connotations, as illustrated by the stative vs.~temporary contrast in \cref{ex:intro5}.

Such distinctions illustrate the general phenomenon of  \textbf{construal}, which we claim has been neglected in NLP. We believe that a proper recognition of construal would provide a unified framework for addressing a wide range of issues involving meaning and linguistic variation% in natural language understanding and generation%
, opening the way to systems that more closely approximate (actually) natural language.
This paper surveys the theoretical and empirical landscape related to construal phenomena and makes the case for its relevance to NLP. After clarifying the terms adopted here (\cref{sec:meaning}), we lay out a few key dimensions of construed meaning (\cref{sec:dimensions}) and then %\textsc{specificity}, \textsc{boundedness}, \textsc{metaphor}, \textsc{metonymy}, \textsc{prominence}, and \textsc{perspective}.
elaborate on some mechanisms of construal (\cref{sec:ops}). %\emph{Construal operations} such as coercion add flexibility 
A trio of case studies illustrate how different types of construal can challenge NLP systems (\cref{sec:casestudies}). We end with some conclusions and suggestions for how to begin addressing these challenges (\cref{sec:discussion}).
%While there is NLP work addressing some aspects of construal, e.g.~metaphor, we contend that the broader issue of construal is a fundamental challenge for the field.

%The goal of this paper is: a) to provide a review of theoretical and empirical work investigating event construal, and b) to argue that current approaches to NLP could benefit from an appreciation of these dimensions of meaning, both in terms of downstream applications (e.g., opinion mining, summarization, machine translation) and in the creation of new, diverse benchmarks with which to evaluate our models.

\section{Meaning and construal}\label{sec:meaning}
%Linguists and philosophers have long recognized the flexible, subjective, and contextually grounded nature of meaning.

Our view of \textit{construal} and its close companion \textit{meaning} is rooted in both frame-based and cognitive semantic traditions. % \st{Flagging this first sentence as a candidate to cut if we're short on space.}
The notion that words and other linguistic units evoke background scenes along with specific perspectives on those scenes is captured by \citeposs{TheCaseforCaseReopened} slogan, \textsc{meanings are relativized to scenes}. This idea has deeper consequences than merely assigning different semantic roles to examples like \cref{ex:intro2}. As \Citet[p.~460]{langacker1993universals} observes, ``any given situation can be viewed in multiple if not infinitely many ways. Starting from the same basic conceptual content\ldots we can form an endless variety of specific conceptions by making alternate choices in regard to the many dimensions of construal.''

%More generally, speakers make many lexical, grammatical and other expressive choices that can preserve truth-conditional semantics but nevertheless shape the comprehender's understanding.

% Philosophers and linguists alike have wrestled for decades with the challenge of characterizing the relationship between language, speakers, and the world. Many have converged on the argument that a purely truth-conditional framework is inadequate for semantics. 

% The potential for linguistic choices to shape understanding---crucially, without affecting anything akin to truth conditions---undermines the idea of a single objective interpretation. ``The meaning of a linguistic expression is more than an objective or universal level of conceptual representation. The choice of words and syntactic structures reflects a conceptualization or construal of the experience being communicated by the speaker'' \citep[p.~52]{croft-00}.\ttor{Isn't there some formatting guideline about indenting long citations?} \st{Could maybe remove or shorten this paragraph, I think it's somewhat redundant with the previous and following.}

This view of linguistic meaning---which we might call inherently \textit{multivalent}---is more flexible than in many theoretical and computational treatments, particularly truth-conditional approaches that liken meanings to facts in a database. The visual domain offers a more informative analog: a photographic or artistic rendering of a scene can vary in vantage point, viewing distance, objects in sight or in focus, color and lighting choices, etc.\ \citep{langacker1993universals, talmy2006grammatical}. Context matters, too: a painting hanging on a preschool wall may be received differently if displayed in a museum. Just as there is no one objective, context-independent depiction of a scene, there are many valid ways to present an idea through language.% \st{I think this paragraph conveys the core idea really well!}

We thus extend Fillmore's slogan to include all kinds of conceptual content (beyond scenes); the broader communicative context; and the effect of choices made as part of the construal process:
\begin{center}\textsc{meanings are relativized to content, context and construal.}\end{center}

Below we elaborate on how each of these interrelated factors affects construed meaning.

\paragraph{Conceptual content.} We assume that linguistic units can evoke and combine all kinds of conceptual content, including open-ended world knowledge (entities, actions, events, relations, etc.)\ as well as more schematic structures often associated with grammar and function words. Crucially, concepts must also be amenable to certain kinds of transformation (e.g., shifts in perspective or granularity) as part of construal; see below.%
\footnote{%
We are not here concerned with precisely how concepts are represented or learned, since we believe the insights related to construal apply broadly across theoretical frameworks.}% See discussion for more.}
%. embodied, experiential basis of meaning \nc{add cites}. See also \citeposs{tamari-20} treatment directed at an NLP audience.
%, or how universal, domain-general or language-specific they may be. 

%Semantic frames are one natural candidate especially suited for capturing perspective shifts among event participants; inheritance but other repositories are possible (e.g. AMR-style predicates, embeddings, etc.). 

%These include attentional phenomena (highlighting some event participants over others, or taking a specific perspective) \citep{fillmore82:_frame}); structuring information about how concepts from one domain are conceptualized in terms of another (metaphor) or described at multiple levels of granularity.

\paragraph{Communicative context.} We take meaning to encompass scene-level entities and events, discourse-level information about the interlocutors and their communicative intents, and other phenomena straddling the (fuzzy) semantic-pragmatic boundary, related to attention (e.g., profiling and perspective) and conditions of usage falling under what
\Citet{fillmore-85} dubbed ``U-Semantics'' (in contrast to truth-oriented ``T-Semantics'').%
\footnote{For example, only U-Semantics can explain why ``the children are \emph{on} the bus'' is preferred over ``the children are \emph{in} the bus'' if the bus is in transit, despite referring to the same spatial relationship.}

Contextual factors (e.g., the interlocutors' identity, beliefs, goals, conceptual repertoire, cultural backgrounds) can radically alter construed meaning\considercutting{---and are themselves subject to construal. How we understand \textit{colorless green ideas} may hinge on whether the speaker is (or is seen as) a theoretical linguist.}. On this view, meaning is not arbitrarily subjective, or merely intersubjective; it is also constrained by all aspects of the communicative context.

\paragraph{Construal.} We define \textbf{construal} as a dynamic process of \textit{meaning construction}, in which speakers and hearers encode and decode, respectively, some intended meaning in a given communicative context. To do so, they draw on their repertoire of linguistic and conceptual structures, composing and transforming them to build coherent interpretations consistent with the speaker's lexical, grammatical, and other expressive choices.\footnote{Both speakers and hearers engage in construal: speakers, in choosing how to present the idea, experience or other content they wish to convey; hearers, in reconstructing that intended meaning. Words like `analysis' and `interpretation' should thus be understood as applying to meaning construction by either interlocutor. (We do not focus here on the many differences between comprehension and production.)}

We take construal to be fundamental to all language use, though how \emph{much} construal and what \emph{kinds} of construal vary across interpretations.%
\footnote{Conventionality plays an important role here: initially creative expressions may require less construal as they become entrenched and their meanings more efficiently accessed.}
In the simplest cases, the relevant components fit neatly together (\`{a} la compositional semantics). But many (or even most) utterances involve a myriad of disparate structures---conceptual, linguistic, and contextual---that may need to be transformed, (re)categorized, or otherwise massaged to be integrated into a single coherent whole.

This conceptual flexibility is not arbitrary: the space of combinatorial options is delimited by \textbf{construal operations} defined with respect to certain privileged \textbf{construal dimensions}. A number of dimensions and operations have been proposed, many motivated by general cognitive processes; we will review some of these in \cref{sec:dimensions}, and illustrate how they are engaged during language use in \cref{sec:ops}.

This inclusive, flexible view of meaning has broad implications for a wide variety of linguistic phenomena, and many parallels in prior work---far too many to address exhaustively here. We restrict our current scope in several ways: (1) While some aspects of context will be mentioned below, we do not address many phenomena related to pragmatic inference (e.g.~politeness, indirect requests). (2) Though many construal dimensions are relevant cross-linguistically, we will not address typological patterns in the lexical, grammatical, and cultural conventions that influence construal. (3) We highlight construal phenomena that are psycholinguistically attested and\slash or relevant to NLP research.

% MOVE SPR to somewhere later. The maximally inclusive, flexible view of meaning presented here has many precedents, besides those noted earlier. The idea that multiple dimensions of meaning may be associated with a linguistic pattern can be seen as a generalization of \Citeposs{dowty1991thematic} \textit{semantic proto-roles}---which link the subject\slash object asymmetry to clusters of agent-like or patient-like properties, respectively. 
% USE SOMEWHERE? Operations that systematically manipulate meaning are reminiscent of \textit{lexical rules} accounting for verbal alternations or polysemy, as well as \textit{type coercions} accounting for aspectual shifts \cite{moens-88}. \nc{add vendler, inter alia}. Additional connections to prior theoretical and empirical work will be mentioned below.

% also connections to: generalization, polysemy, metonymy, metaphor, schematicity/embodiement, non-compositionality.

%\nss{I think the intention is that construal is required to interpret content and context.}

%comments apply to both explicit/inferred information, static/dynamic meaning

%\nss{Mention that \citet{bender-20} acknowledge content and context but not construal}

\section{Dimensions of construed meaning}\label{sec:dimensions}
Several (partial) taxonomies of construal dimensions have been proposed in the cognitive linguistics literature \citep{langacker1993universals, talmy2006grammatical, croft-00, taylor1995introduction, casad1995seeing}; see \citet{croft-cruse-2004} for an overview. We will not attempt to reconcile their many differences in terminology and organization, but instead present selected dimensions most relevant for NLP. %These fall into \shortlong{2}{two} broad categories: \textsc{perspective}, \textsc{prominence}, and \textsc{resolution} are all concerned with how concepts are \textit{perceived}, while \textsc{configuration} and \textsc{metaphor} involve the choice or \textit{categorization} of relevant domains and concepts. 

%and \textsc{Schematization}

\subsection{Perspective} \label{sec:perspective}

Languages have many ways of describing scenes from a specific \textsc{perspective} (or \textit{vantage point}). The spatial domain provides clear examples: a cup might be described as being \textit{left} or \textit{right} of some other object, depending on whose perspective is adopted; or explicitly marked as being \textit{on my\slash your\slash her\slash Sue's left}. Likewise, the same motion event can be described relative to differing deictic centers (e.g., the \textit{arrival} in \cref{ex:nora} can also be viewed as a departure from the hospital).

Perspective can extend beyond the spatial domain. The use of past tense in \cref{ex:nora} indicates the speaker's retrospective viewpoint. Differences in opinion, belief state or background have also been treated as perspective shifting.\considercutting{\footnote{See also work on the cognitive dynamics of \emph{perspective-taking} \cite{dale2018interacting, brown2011talking}.} }

%\longversion{Note that this is distinct from the issue of which element is \emph{profiled} or made prominent (see \cref{sec:prominence} above); in both examples just given, ``the cup'' corresponds to the Trajector, and ``the plate'' corresponds to the Landmark, but they differ in which vantage point the speaker adopts or intends the comprehender to adopt.}

%\longversion{Vantage point need not be identical to the speaker's egocentric perspective. A speaker can adopt the perspective of one of the participants in a scene (e.g., ``The cup was to \emph{her left}'') or of the interlocutor (e.g., ``The cup is to \emph{your left}'').}

 \citeposs{talmy2006grammatical} taxonomy defines a broader version of \textsc{perspective} that includes distribution of attention. Descriptions of a static scene can adopt a dynamic perspective, evoking the experience of moving through the scene (``There is a house every now and then through the valley''); these descriptions can be even more explicit, as with \textbf{fictive motion} (``The road runs through the valley'') \citep{talmy1996fictive, matlock2004fictive}.

\paragraph{Psycholinguistic evidence.}
% TEMP REMOVE: Fictive motion can influence the way comprehenders conceptualize a static scene \citep{matlock2004conceptual, matlock2004fictive}, leading them to simulate traversing or scanning the scene in a direction implied by the sentence \citep{matlock2004conceptual, matlock2004fictive, ramscar2010time, matlock2004eye}. 

% Furthermore, various aspects of this simulation appear to be sensitive to details of the scene; for example, eye movements scanning a path described by a sentence (e.g., ``The road goes through the desert'') were slower when the terrain was described as difficult (``The desert is hilly'') vs. easy to traverse (``The desert is flat'') \citep{richardson2007integration}. Even more strikingly, the presence of fictive motion can affect how participants interpret temporally ambiguous statements like ``move the meeting forward'' \citep{matlock2005experiential}, further solidifying the conceptual link between our understanding of space and time (see \cref{sec:metaphor} on \textsc{metaphor}).

Grammatical person can affect which perspective a comprehender adopts when reading about an event \citep{brunye2009you} and which actions they are most likely to remember \citep{ditman2010simulating}. Fictive motion can also influence the way comprehenders conceptualize a static scene \citep{matlock2004conceptual, matlock2004fictive}.

%\textbf{Grammatical person} (\textit{I}\slash \textit{you}\slash \textit{they} etc.) is perhaps the most obvious correlate to this dimension; 

\paragraph{Relevant NLP research.} Perspective is crucial for understanding spatial language, e.g.~for robotics (\cref{sec:robotics}) and other kinds of situated language. Work on grounding referents from natural language descriptions has incorporated visual perspective as another source of information about the intended referent \cite{devin2016implemented, ros2010one, trafton2005enabling}.

\subsection{Prominence} \label{sec:prominence}

\textsc{prominence} (or \textit{salience}) refers to the relative attention focused on different elements in a scene \citep{langacker1993universals,talmy2006grammatical}. Languages have various devices for highlighting, or \emph{profiling}, some elements over others (or leaving them implicit). For example, verbs like those in \cref{ex:intro2} differ in which elements in a larger scene are preferentially expressed. Similarly, many spatial and temporal adpositions involve an asymmetric profiling of one entity relative to another; thus ``the painting is above the piano'' and ``the piano is below the painting'' describe the same situation but differ in focus.

Verbal and constructional alternations also manipulate prominence: The active/passive pair ``Microsoft employs me'' and ``I am employed by Microsoft'' differ in profiling the employer and speaker, respectively. Similarly, transitive ``I rolled the ball'' vs.~intransitive ``The ball rolled'' differ in whether the ball-roller is even mentioned.

Languages also differ systematically in how motion events are most idiomatically expressed, in particular in whether the main verb encodes (and foregrounds) the manner (English \textit{run}) or path (Spanish \textit{entrar}) of motion. %\nc{ed; move to fn?}

\paragraph{Psycholinguistic evidence.}

A speaker's decisions about which features to encode in the main verb versus a satellite can influence which events comprehenders find most similar \citep{billman1998path} and which features they tend to remember \citep{gennari2002motion}.

% \nc{This example is a bit long and requires verb-satellite background. Can we find a different example }\nss{agreed. also seemed jarring to suddenly give examples in Spanish as well as English when we said typological variation wouldn't be our focus}\st{Cut this down and added a description of the phenomenon in the main section. Keeping for now b/c there's also intra-lingiustic variation in construal---will cut if need space.}

In other work, \citet{fausey2010subtle} found that descriptions of an accidental event using a transitive construction (``She had ignited the napkin'') led participants to assign more blame to the actor involved, and even demand higher financial penalties, than descriptions using non-agentive constructions (``The napkin had ignited'').

In language production, there are a number of factors influencing which construction a speaker chooses (e.g., current items in discourse focus \citep{bresnan2007predicting}, lexical and syntactic priming \citep{pickering2008structural}).

\paragraph{Relevant NLP research.}

% Prominence is a very broad topic, covering what is explicit vs.~implicit and emphasized vs.~backgrounded. 
Recovering implicit information is widely studied in NLP, and deciding which information to express is key to NLG and summarization. We mention three examples exploring how choices of form lend prominence to certain facets of meaning in ways that strongly resonate with our claims about construal.

%\nss{added this sentence to clarify why these particular choices}

\Citet{greene-09} show that syntactic framing---e.g.~active (\emph{Prisoner murders guard}) vs.~passive (\emph{Guard is murdered})---is relevant to detecting speaker sentiment about violent events. %\nc{Yikes, is there a less terrible example we can use?!}\st{+1 to Nancy's point---haven't read the paper but there's got to be other active/passive cases with differing speaker sentiment...}\nss{The dataset was specifically acts of violence in the news...some of the other passages are even more gruesome :(}\nc{but maybe less gendered}

\Citet{hwang-17} present an annotation scheme for capturing adpositional meaning construal (as in \cref{ex:msft-prep}). Rather than disambiguate the adposition with a single label, they separately annotate an adposition's role with respect to a scene (e.g.~employment) and the aspect of meaning brought into prominence by the adposition itself (e.g., benefactive
 \emph{for} vs.~locative \textit{at}).
This more flexibly accounts for meaning extensions and resolves some annotator difficulties.
%, including locative-to-goal coercions with \emph{put}.
%, and can be used to characterize crosslinguistic divergences in adposition use \citep[cf.][]{chinesesnacs}.
%\Citet{shalev-19} argue that the framework can be expanded to include subjects and objects as well.

%Prominence is also relevant to discourse structure: 

\Citet{rohde-18} studied the construction of discourse coherence by asking participants to insert a conjunction (\lex{and}, \lex{or}, \lex{but}, \lex{so}, \lex{because}, \lex{before}) where none was originally present, before an explicit discourse adverbial (e.g.~\lex{in other words}). They found that some contexts licensed multiple alternative conjunctions, each expressing a different coherence relation---i.e., distinct implicit relations can be inferred from the same passage. This speaks to the challenge of fully annotating discourse coherence relations and underscores the role of both linguistic and contextual cues in coherence.
% Presumably, speakers are forced to choose at most one of multiple coherence relations to highlight in a complex situation with a conjunction, highlighting that aspect of the meaning. 
% \nc{Who is presuming. this? Is this our speculation or what they concluded?\nss{both :)} This also sounds more psycholinguistic than NLPish.\nss{it was a psycholing+CL collaboration, published at ACL}}

\subsection{Resolution}

Concepts can be described at many levels of \textsc{resolution}---from highly detailed to more schematic. We include here both \textit{specificity} (e.g., \textit{pug} $<$ \textit{dog} $<$ \textit{animal} $<$ \textit{being}) and \textit{granularity} (e.g., viewing a forest at the level of individual leaves vs. branches vs. trees). 
%, actions (\textit{march} $<$ \textit{walk} $<$ \textit{move}) and other concepts are arranged with more concrete entries at lower hierarchical levels.
Lexical items and larger expressions can evoke and combine concepts at varying levels of detail (``The gymnast triumphantly landed upright'' vs.~``A person did something'').

\paragraph{Psycholinguistic evidence.}
Resolution is related to \textbf{basic-level categories} \citep{rosch2004basic, lakoff1987categories, hajibayova2013basic}, the most culturally and cognitively salient levels of a folk taxonomy. Speakers tend to use basic-level terms for reference (e.g., \textit{tree} vs.~\textit{entity}\slash \textit{birch}), and basic-level categories are more easily and quickly accessed by comprehenders \citep{mervis1981categorization, rosch2004basic}. 

Importantly, however, what counts as basic-level depends on the speaker's domain expertise \citep{tanaka1991object}. Speakers may deviate from basic-level terms under certain circumstances, e.g., when a more specific term is needed for disambiguation \citep{graf2016animal}. Conceptualization is thus a flexible process that varies across both individual cognizers (e.g., as a function of their world knowledge) and specific communicative contexts. 

\paragraph{Relevant NLP research.}
Resolution is already recognized as important for applications such as text summarization and dialogue generation \citep{louis2012corpus, li2015fast, ko2019domain, li2016improving, ko2019linguistically}, e.g., in improving human judgments of informativity and relevance \citep{ko2019linguistically}. Also relevant is work on knowledge representation in the form of inheritance-based ontologies and lexica (e.g., FrameNet \cite{framenet}, ConceptNet \cite{liu2004conceptnet}). 
%\nc{Do either the psych or CS evidence here address non-inheritance-based granularity, i.e. step vs. walk?}

\subsection{Configuration} \label{sec:config}

\textsc{configuration} refers to internal-structural properties of entities, groups of entities, and events, indicating their schematic ``shape'' and ``texture'':
multiplicity (or \emph{plexity}), homogeneity, boundedness, part-whole relations, etc.\ \citep{langacker1993universals, talmy2000a}.
To borrow an example from \citet{croft-12}, a visitor to New England can describe stunning autumn \textit{leaves} or \textit{foliage}. Though both words indicate a multiplex perception, they exhibit a grammatical difference: the (plural) \textbf{count} noun \textit{leaves} suggests articulated boundaries of multiple individuals, whereas the \textbf{mass} noun \textit{foliage} suggests a more impressionistic, homogeneous rendering. %Nonetheless, both indicate a multiplex perception of the referent.

This dimension includes many distinctions and phenomena related to \textbf{aspect} \cite{vendler-67,comrie1976aspect}, including whether an event is seen as discrete (\textit{sneeze}) or continuous (\textit{read}); involves a change of state (\textit{leave} vs.~\textit{have}); has a defined endpoint (\textit{read} vs.~\textit{read a book}); etc. 
Lexical and grammatical markers of configuration properties interact in complex ways; see discussion of 
count\slash mass and aspectual coercion in \cref{sec:ops}.

\paragraph{Psycholinguistic evidence.}

Differences in grammatical aspect can modulate how events are conceptualized \citep{matlock2011conceptual}. Stories written in imperfective aspect are remembered better; participants are also more likely to believe that the events in these stories are still happening \citep{magliano2000verb} and build richer mental simulations of these events \citep{bergen2010grammatical}. In turn, these differences in conceptualization have downstream consequences, ranging from judgments about an event's complexity \citep{wampler2019} to predictions about the consequences of a political candidate's behavior on reelection \citep{fausey2011can}. 
%\nc{TODO: imperfective reference -- though one wrinkle is that the imperfective might be more about internal perspective. It does connect to homogeneity but that's probably not the key thing for these psych results.}

The mass/count distinction has attested psychological implications, including differences in word recognition time \cite{gillon1999mass} (see \citet{fieder2014representation} for a review). 

\paragraph{Relevant NLP research.}  
Configurational properties are closely linked to well-studied challenges at the syntax-semantic interface, in particular nominal and aspectual \textit{coercion} effects (\cref{sec:ops}). Several approaches explicitly model coercion operations based on event structure representations \cite{moens-88,passonneau-88,pulman-97,chang-98}, while others explore statistical learning of aspectual classes and features \cite{siegel-00,mathew-09,friedrich-14}. Lexical resources have also been developed for aspectual annotation \cite{donatelli-18} and the count\slash mass distinction \citep{schiehlen-06,kiss-17}.

%the subject of a SemEval task \cite{pustejovsky-jezek} and 
%special issue \cite{aspect-cl-88}
% \citep[e.g.,][]%
%{moens-88,passonneau-88,chang-98,siegel-00,mathew-09,friedrich-14,loaiciga-16,friedrich-17,donatelli-18}.
%\citep{schiehlen-06,kiss-16,kiss-17,smith-17,pustejovsky-jezek}.

\subsection{Metaphor}\label{sec:metaphor}
The dimension of \textsc{metaphor} is broadly concerned with \textit{cross-domain comparison}, in which speakers ``conceptualize two distinct structures in relation to one another'' \citep[p.~450]{langacker1993universals}. Metaphors have been analyzed as structured mappings that allow a \emph{target} domain to be conceptualized in terms of a \emph{source} domain \cite{lakoff-80}.

Metaphors pervade language use, and exhibit highly systematic, extensible structure. For example, in English, events are often construed either as locations in space or as objects moving through space. Our experience of time is thus often described in terms of either motion toward future events (``we're approaching the end of the year''), or the future moving toward us (``the deadline is barreling towards us'')
\citep{boroditsky2000metaphoric, boroditsky2001does, hendricks2017new, nunez2006future}. Metaphor plays a role in our linguistic characterization of many other domains as well \cite{lakoff-80}. 
% , such as mathematics \citep{marghetis2013motion, lakoff2000mathematics} and even musical pitch \citep{dolscheid2013thickness} (see \citet{thibodeau2017linguistic} for a review).

\paragraph{Psycholinguistic evidence.}
Different metaphors can shape a comprehender's representation about the same event or concept in radically different ways. \citet{thibodeau2011metaphors} found that describing a city's crime problem as a \textit{beast} or as a \textit{virus} elicited markedly different suggestions about how best to address the problem, e.g., whether participants tended to endorse enforcement- or reform-based solutions. Similar effects of metaphor on event conceptualization have been found across other domains, such as cancer \citep{hauser2015war, hendricks2018emotional} and climate change \citep{flusberg2017metaphors} (see \citet{thibodeau2017linguistic} for a thorough review). 
%Moreover, as with the other dimensions of meaning discussed here, the effect of metaphor framing is known to vary across individuals \citep{elmore2017light, stamenkovic2019metaphor}.

\paragraph{Relevant NLP research.}
Considerable NLP work has addressed the challenge of metaphor detection and understanding \citep{narayanan-00,shutova2010metaphor, shutova2013statistical, shutova2015design}. This work has made use of both statistical, bottom-up approaches to language modeling \citep{gutierrez2016literal, shutova2013statistical}, as well as knowledge bases such as MetaNet \citep{dodge2015metanet, stickles2014construction, david2017computational}.

\subsection{Summary}

%We have presented several dimensions along which meaning can be systematically construed, along with psycholinguistic evidence for the cognitive and communicative importance of each dimension; we have also provided a brief review of NLP research that has tackled phenomena relevant to each dimension. 

The selective review of construal dimensions presented here is intended to be illustrative, not exhaustive or definitive. 
Returning to the visual analogy, we can see these dimensions as primarily concerned with how (and what part of) a conceptual ``scene'' is \textit{perceived} (\textsc{perspective}, \textsc{prominence}); the choice or \textit{categorization} of which schematic structures are present (\textsc{configuration} and \textsc{metaphor}); or both (\textsc{resolution}).

We have omitted another high-level categorization dimension, \textsc{schematization}, which includes concepts related to force dynamics, image schemas, and other experientially grounded schemas well discussed in the literature \cite{talmy2000a}. We have also not addressed pragmatic inference related to politeness \citep{brown1987politeness}, indirect requests \citep{clark1979responding}, and other aspects of communicative intent. Additionally, some phenomena are challenging to categorize within the dimensions listed here; a more complete analysis would include evidentality \cite{chafe1986evidentiality}, modality \cite{mortelmans2007modality}, light verb constructions \cite{wittenberg2017if, wittenberg2014processing}, and more. Nonetheless, we hope this partial taxonomy provides a helpful entry point to relevant prior work and starting point for further alignment.
%These fall into \shortlong{2}{two} broad categories: \textsc{perspective}, \textsc{prominence}, and \textsc{resolution} are all concerned with how concepts are \textit{perceived}, while \textsc{configuration} and \textsc{metaphor} involve the choice or \textit{categorization} of relevant domains and concepts. 

% \nc{modified above. need more cites for embodiment section?}

%In section \cref{sec:ops}, we now turn to well-documented mechanisms by which construal can occur in real-time language use.

\section{Construal in action}\label{sec:ops}
How might construal work in practice? We have emphasized so far the \emph{flexibility} afforded by the dimensions in \cref{sec:dimensions}. But we must also explain why some words and concepts make easier bedfellows than others. This section presents a thumbnail sketch of how the construal process copes with apparent mismatches, where it is the collective \emph{constraints} of the input structures that guide the search for coherence.

%\footnote{Note that some of the phenomena discussed here (e.g., coercion) have treatments in other approaches, given that they can affect the truth conditions of an utterance. 
%We discuss them to emphasize how the construal dimensions and operations offer a unified conceptual basis for why apparently mismatched elements can combine.\nss{OK? Nancy, this arose from a discussion after you left over the hypothetical reader's confusion that in this section we are suddenly talking about phenomena that DO affect truth conditions}
%Nonetheless, a holistic appreciation of construal has the advantage of, within one same framework, both accounting for truth condition semantics and allowing for the the explanation of important aspects of meaning that are beyond truth conditions.}

%cues present during language use can \emph{trigger} construal operations in the search for coherence.

We focus on comprehension (similar processes apply in production), and assume some mechanism for proposing interpretations consisting of a set of conceptual structures and associated \textbf{compatibility constraints}. Compatibility constraints are analogous to various kinds of binding constraints proposed in the literature (variable binding, role-filler bindings, unification bindings, and the like): they are indicators that two structures should be conceptualized as a single unit. But compatibility is softer and more permissive than identity or type-compatibility, in that it can also be satisfied with the help of \textit{construal operations}. Some operations effect relatively subtle shifts in meaning; others have more dramatic effects, including changes to truth-conditional aspects of meaning.
%\nss{How about: \textbf{Construal operations} may apply to restructure the input structures until compatibility constraints are satisfied and a coherent interpretation is formed.}

% An interpretation fails (or is deemed unlikely or low in coherence) if no sequence of operations satisfies all compatibility constraints.

%Below, we list several examples of these \textbf{construal operations}, but they are discussed in considerably more detail elsewhere \citep{talmy1977rubber, talmy2006grammatical, moens-88}. 

Below we illustrate how some example linguistic phenomena fit into the sketch just presented and mention connections to prior lines of work.

%\nss{count/mass coercion? ``a water'', ``cat all over the driveway''. but not unlimited: *furnitures. seems a simpler-to-understand starting point that would lead into aspect} \ttor{Indeed, easier, but, already mentioned before, so maybe readers would already be familiar with that?}

%of their default \textsc{configuration} property,
\paragraph{Count/mass coercion.} English nouns are flexible in their \textbf{count}\slash \textbf{mass} status (see \cref{sec:config}). Atypical marking for number or definiteness can cause a shift, or \textbf{coercion}, in boundedness: plural or indefinite marking on mass nouns (\textit{a lemonade}, \textit{two lemonades}) yields a bounded interpretation (cups or bottles of lemonade). Conversely, count nouns with no determiner are coerced to an undifferentiated mass, via a phenomenon known as \textbf{grinding} (``there was mosquito all over the windshield'') \citep{pelletier-89,pelletier-03,copestake-95}.
Here we see evidence of the outsize influence of tiny grammatical markers on manipulating lexical defaults in the construal process.

%%\cref{sec:config}
\paragraph{Aspectual composition.} 
Aspect is a prime arena for studying how multiple factors conspire to shape event construal. Verbs are associated with default aspectual classes that can be coerced under pressure from conflicting cues, where details of event structure systematically constrain possible coercions and their inferential consequences \cite{moens-88,talmy2006grammatical}.
 
%grammatical markers, arguments and modifiers, and context
%Aspect is a prime arena for studying how conflicting constraints---in this case, on how an event unfolds over time---may trigger operations that systematically transform meaning \cite{moens-88,talmy2006grammatical}. % We assume construal operations are essentially coercion operations, but with explicit links to construal dimensions. \st{Not sure this is needed. Cutting in the interest of space.}

In fact, aspectual coercion can be reanalyzed in terms of construal dimensions. For example, \textit{durative} modifiers (e.g.~\textit{for an hour}) prefer to combine with
\textbf{atelic} processes (lacking a defined endpoint, as in \ref{ex:asp1}) on which to impose a bound (analogous to count/mass coercion) and duration. Combination with any other aspectual class triggers different operations to satisfy that preference:
%of this work; \cite{chang-98} extends this analysis to a richer event structure model.

\ex.\label{ex:aspect} 
    \a.\label{ex:asp1} He \{slept / ran\} for an hour.
    \b.\label{ex:asp2} He sneezed for an hour. 
    \b.\label{ex:asp3} He read the book for an hour.
    \b.\label{ex:asp4} He left for an hour.
    \z.

A single sneeze, being a discrete event unlikely to last an hour, undergoes \textsc{iteration} into a series of sneezes \cref{ex:asp2}, illustrating a change in plexity (\cref{sec:config}); while the book-reading in 
in \cref{ex:asp3} is simply viewed as unfinished (cf.~``He read the book''). The departure in \cref{ex:asp4} is a discrete event, but unlike sneezing, it also results in a state change that is reversible and therefore boundable (cf.~the iterative reading of ``He broke the glass for an hour'', the non-permanent reading of \ref{ex:intro5}). Its coercion thus features multiple operations: a \textsc{prominence} shift to profile the result state of being gone; and then a \textsc{bounding} that  also reverses state, implying a return \cite{chang-98}.

\paragraph{Constructional coercion.} The flagship example cited in the construction grammar literature \cref{ex:sn1} has also been analyzed as a kind of coercion, serving to resolve conflicts between lexical and grammatical meaning\longversion{ (with grammatical meaning taking precedence)} \citep{goldberg1995constructions, goldberg2019explain}:

\ex.\label{ex:sneeze} 
    \a.\label{ex:sn1} She sneezed the napkin off the table.
    \b.\label{ex:sn2} She \{pushed / blew / sneezed / ?slept\} the napkin off the table.
    \z.

Here, the verb \textit{sneeze}, though not typically transitive or causal, appears in a \textbf{Caused Motion} argument structure construction, which pairs oblique-transitive syntax with a caused motion scene. The resulting conflict between its conventional meaning and its putative causal role is resolvable, however, by a commonsense inference that sneezing expels air, which can plausibly cause the napkin's motion \citep[cf.][]{forbes-17}.

This coercion, also described as \textit{role fusion}, differs from the previous examples in manipulating the \textsc{prominence} of a latent component of meaning. 
Coercion doesn't always succeed, however: presumably sneezing could only move a boulder with contextual support, and  sleeping has a less plausibly forceful reading. In fact, construal depends on the interaction of many factors, including degree of conventionality (where \textit{push} and \textit{blow} are prototypical caused motion verbs), embodied and world knowledge (the relative forces of \textit{sneeze} and \textit{sleep} to napkin weight), and context.%
\footnote{A related theory is \citeposs{dowty1991thematic} \textit{semantic proto-roles} account, which links the grammatical subject\slash object asymmetry to two clusters of semantic features that are more agent-like (e.g., animacy) or patient-like (e.g., affectedness), respectively; associations between these \textit{proto-roles} and grammatical subjects and objects are attested in comprehension \citep{kako2006thematic, pyykkonen2010three} and have been investigated computationally \cite{reisinger2015semantic, rudinger-18}.}

There is extensive psycholinguistic evidence of constructional coercion and the many factors influencing ease of construal (see \citet{goldberg2003constructions, goldberg2019explain} for reviews). Some of these phenomena have been analyzed within computational implementations of construction grammar  \cite{bergen-chang-05,bryant-08,bergen2013embodied,dodge2014representing,steels-17-fcg,comp-cxg-aaai-18,SSS1715257}, and have also been incorporated in corpus annotation schemes \cite{bonial-11, hwang-14, lyngfeltetal2018}.

\paragraph{Metonymy and metaphor.} Metonymy and metaphor are associated with semantic mismatches that trigger construal operations. A possible analysis of \longversion{the phrase }\textit{tiny iron grip} from \cref{ex:nora} illustrates both.

First, the modifiers \textit{tiny} and \textit{iron} expect a physical entity, but \textit{grip} is a (nominalized) action. This conflict triggers a profile shift (\textsc{prominence}) to the grip's effector (a hand), effectively licensing a metonymy. A further conflict arises between the hand and its description as \textit{iron} (unlikely to be literal unless the protagonist is of robotic lineage). A structural alignment (\textsc{metaphor}) then maps the iron's strength to the grip's force, which in turn maps to the degree of dictatorial control.%
\footnote{Alternatively, \textit{iron grip} could be treated as an entrenched idiom with a readily accessible construal that \textit{tiny} can modify.}

%\footnote{Conventionality plays an important role here: initially creative expressions may require less construal as they become entrenched and their meanings more efficiently accessed.}

We observe that multiple construal operations can occur in sequence; that a conceptual or linguistic element may afford more than one construal within the same analysis (\textit{grip} as both a hand and metaphorical control); and that aspects of common sense, world knowledge, and culture (though not the focus of the present work) inevitably constrain construal options.

\noindent %Note that both sentences feature the same verb, \emph{quebrar}, and the same syntactic structure. Nonetheless, 

% Despite the superficial similarity, there is a key difference: the car causes the breaking in \cref{ex:carma_nut}, but not in \cref{ex:carma_engine}. %\nss{Is it fair to say that this is disambiguated by recognizing that a screen cannot be an agent (it is inanimate), or is there more to it?}\ttor{Actually, there is. The same construction can be used with human subjects, in examples such as "A menina quebrou a perna" 'The girl broke the/her leg' and "Lewis Hamilton quebrou o motor" 'The engine in Lewis Hamilton's car failed.'}  \Citet{SSS1715257} show how such constructions can be disambiguated automatically using a rich constructicon resource.
%In the system proposed by \citet{SSS1715257}, spreading activation techiniques are used on a graph representation of the Brazilian Portuguese Constructicon, a structured repository of constructions for that language, so as to generate the correct semantic representation for each sentence. 

\section{Case studies}\label{sec:casestudies}
We turn to a few illustrations of how the pervasive effects of construal can arise in applied settings.

\subsection{Case study 1: Conversational assistants}\label{sec:dialog}

Even simple tasks like rescheduling a meeting pose many challenges to dialogue systems, in both understanding users' intents and formulating natural responses. Consider the following exchange:

\begin{dialogue}
\speak{U-1} When is my 1-1 with Chuck?
\speak{A-2} 4 PM today, in 15 minutes.
\speak{U-3} Is there another slot soon?
\speak{A-4} Not today, should I check tomorrow?
\speak{U-5} Let's push it to his tomorrow evening.
\speak{A-6} Rescheduled 1-1 with Chuck for 2 PM tomorrow, 6 PM in Brazil.
\end{dialogue}

The agent's first response (A-2) demonstrates sensitivity to \textsc{perspective} by providing a relative time. Interpreting ``another slot soon'' in the user's follow-up (U-3) requires both understanding that \textit{another} is implicitly defined in contrast to the existing slot (relying on \textsc{prominence}) and then inferring the appropriate \textsc{resolution} meant by \textit{soon} (on the scale of hours, rather than minutes or seconds). The agent's succinct response in (A-4) exploits \textsc{prominence} yet again, both by eliding reference to the sought-after open meeting slot with Chuck, and by using ``tomorrow'' (the direct object of ``check'') as a metonymic shorthand for the joint constraints of the user's and Chuck's calendars.

The next user turn (U-5) employs \textsc{metaphor} in its construal of an event as a physical object, capable of being pushed. The metaphorical destination (``his tomorrow evening'') requires consideration of differing time zones (\textsc{perspective}), as made explicit in the final agent turn (A-6).

Interactions between situational context and the kinds of compatibility constraints discussed in \cref{sec:ops} can also affect a dialogue system's best response. 
A user asking a fitness tracking app ``How long have I been running?''\ while panting around a track may be referring to the current run, but the same question asked while sitting at home is more likely wondering how long they've been \emph{habitually} running.
A successful response requires the integration of the constraints from (at least): the verb \textit{running}, whose progressive marking is associated with ongoing processes, but ambiguous between a single run and a series of runs (\textsc{configuration}); the present-perfect \textit{have been V-ing}, which implies an internal view (\textsc{perspective}); and the situational context (is the user currently running?).

%because a series of events can be construed as a macroevent spanning from the start of the first instance to the end of the last.
%
%expect a short-term answer (i.e., ``How many minutes has it been since I started my current run?''), while a user sitting at home might expect a long-term answer (i.e., ``How long did I begin running for exercise''), measured in days or months. 
%
%If the user who is in the middle of a race asks ``How long have I been running?'', they are likely expecting a short-term answer (i.e., `How many minutes has it been since I started my current run today?').
% By contrast, if this same question is asked by a user who is not on a run, the intention probably was something like `How long ago did I begin running for exercise?', and a long-term answer (measured in days or months) would be appropriate. %\footnote{\emph{Common ground} between the user and the agent is also relevant here: if the user has been running regularly for many years but has only recently started using the fitness app, it would be reasonable to assume that the user expects the agent only knows about this period of time. }
%`How long?'\ can question both the duration of a single event, as well as the duration of a \emph{series} of events---because a series of events can be construed as a macroevent spanning from the start of the first instance to the end of the last.

\subsection{Case study 2: Human-robot interaction}\label{sec:robotics}

Situated interactions between humans and robots require the integration of language with other modalities (e.g., visual or haptic).%
\footnote{Indeed, the needs of human-robot interaction have motivated extensions to Abstract Meaning Representation \citep{amr} beyond predicate-argument structure and entities to capture tense and aspect, spatial information, and speech acts \cite{bonial-19}.}%
\ Clearly, any spatially grounded referring expressions must be tailored to the interlocutors' \textsc{perspective} (whether shared or not) \citep{kunze2017spatial}.

Focus of attention (\textsc{prominence}) is especially important for systems that must interpret procedural language. Recipes, for example, are notoriously telegraphic, with rampant omissions of information that a human cook could easily infer in context \cite{ruppenhofer-10,malmaud-14-cooking}. Consider \cref{excake}:

\ex.\label{excake}\emph{In} a medium bowl, cream \emph{together} the sugar and butter. Beat \emph{in} the eggs, one at a time, then stir \emph{in} the vanilla.

\noindent The italicized words provide crucial constraints that would help a cook (human or robot) track the evolving spatial relations. The first \emph{in} establishes the bowl as the reference point for the creaming action, whose result---the mixture of sugar and butter \emph{together}---becomes the implicit landmark for the subsequent beating \emph{in} of eggs and vanilla.

%Consider first the case of \textsc{resolution}.
%A fundamental problem in robotic learning is linking low-level sensorimotor data to language \citep{7856986}, sometimes called the \emph{symbol grounding problem} \citep{HARNAD1990335}. 

%A system tasked with recognizing, labeling, and performing the sequence of actions involved in some activity (e.g., cooking dinner) must first be able to \emph{segment} the continuum of sensorimotor data in a manner consistent with linguistic descriptions of those actions. Among other things, this mapping requires a system to understand that the same action can be described at varying levels of resolution (e.g., ``cut the carrots'' vs.~``dice the carrots'').

Systems following instructions also require a means of segmenting continuous sensorimotor data and linking it to discrete linguistic categories \citep{regneri-13,yagcioglu-18} (cf.~the \emph{symbol grounding problem} \citep{HARNAD1990335}). This mapping may depend on flexibly adjusting \textsc{resolution} and \textsc{configuration} based on linguistic cues (e.g., \textit{cut/dice/slice/sliver the apple}).

\longversion{Perspective may also extend to non-spatial domains, such as knowledge states. People differ in what they know and don't know about the world, and these differences can affect the interpretation of what they mean by what they say \citep{epley2004perspective, brown2011talking, trott2018individual}; a system could track potential divergences in knowledge states using separate ontologies \citep{lemaignan2010oro}, then use these divergences to augment pragmatic inference \citep{williams2014dempster, trott2017theoretical}.}

%\nss{TODO: Bender Rule check}

\subsection{Case study 3: Paraphrase generation}\label{sec:captioning}

Despite many advances, paraphrase generation systems remain far from human performance. One vexing issue is the lack of 
evaluation metrics that correlate with human judgments for tasks like paraphrase, image captioning, and textual entailment \citep[see, e.g.,][]{bhagat-13,pavlick-19, wang2019task}.

In particular, it is unclear how closely a good paraphrase should hew to \emph{all} aspects of the source sentence. For example, should active\slash passive descriptions of the same scene, or the sets of sentences in \cref{ex:intro}, be considered meaning-equivalent? Or take the putative paraphrase below:
\ex.\label{ex:para}
    \a.\label{ex:para1} The teacher sat on the student's left.
    \b.\label{ex:para2} Next to the children was a mammal.

These could plausibly describe the same scene; should their differences across multiple dimensions (\textsc{perspective}, \textsc{prominence}, \textsc{resolution}) be rewarded or penalized for this diversity?

A first step out of this quandary is to recognize construal dimensions and operations as a source of linguistic variability. Paraphrase generation and other semantically oriented tasks could incorporate these into system design and evaluation in task-specific ways.

%Therefore, the standards for evaluating such systems, as well as the systems themselves, may benefit from 
%isolating or representing different aspects of construal.
%some sort of meaning representation that includes construal.

\section{Discussion}\label{sec:discussion}

Throughout this paper, we have emphasized the flexible and multivalent nature of linguistic meaning, as evidenced by the \textbf{construal} phenomena described here. The effects of construal are ubiquitous: from conventional to creative language use, through morphemes and metaphors. Indeed, even the smallest forms can, like tiny tyrants, exert a transformative force on their surroundings, inducing anything from a subtle shift in emphasis to a radical reconceptualization.

%Throughout this paper, we have emphasized the flexible and multivalent nature of linguistic meaning. Events and entities in the world do not exist in 1-1 correspondence with language---rather, the same event is compatible with many descriptions.

%Indeed, \textbf{construal} is inextricable from meaning. Speakers describing an event face many decisions about how to express it in language, including: which features to foreground (\textsc{prominence}), the appropriate level of \textsc{resolution}, whether to employ \textsc{metaphor}, and more (see \cref{sec:dimensions}). In turn, these subtle differences in linguistic form can induce radical shifts in a comprehender's interpretation.

% Took out: "...is a boon for communication and creative language use". :( 
As illustrated in \cref{sec:casestudies}, this flexibility of language use poses a challenge for NLP practitioners. Yet crucially---and fortunately---construal is not random: \textbf{variations in linguistic form correspond \emph{systematically} to differences in construal}. 
The dimensions of construal and their associated operations (\cref{sec:dimensions} and \cref{sec:ops}) offer principled constraints that render the search for coherence more tractable.
%This systematicity renders the problem of modeling meaning more tractable.

How, then, should we proceed? Our goal is for construal dimensions such as those highlighted in \cref{sec:dimensions} to be incorporated into any research program aspiring to human-level linguistic behavior. Below, we describe several concrete recommendations for how to do this.

%More concretely, insights from construal should help encourage generation systems to produce diverse outputs; improve evaluation systems for natural language inference and translation; and inform the design of linguistic resources and meaning representations. We elaborate on two of these angles below.
%\nc{why not all? just space? seems weird to mention and then say nothing more about 3rd.}\nss{oh, I didn't read the ``More concretely'' sentence as a list of 3 main goals, I read it as a few examples. But yeah, it doesn't quite fit the rest of the section. Can it be removed?} \st{Yes, removed and simplified it.}
%In the section below, we discuss several concrete approaches that we believe would be fruitful.

%\subsection{Looking forward: recommendations}
%Concretely, we suggest the following.

% Took out: "..., as well as the most appropriate representational format and implementation of its associated construal operations,..."

%\paragraph{Evaluating the evaluations.}
\paragraph{More meaningful metrics.} Taking construal seriously means rethinking how NLP tasks are designed and evaluated. Construal dimensions can provide a rubric for assessing tasks, datasets, and meaning representations \citep{abend-17} for which meaningful distinctions they make or require. (E.g.: Does it capture the level of \textsc{resolution} at which entities and events are described? Does it represent \textsc{metaphor}? Is it sensitive to the \textsc{prominence} of different event participants?)

Such questions might also help guard against unintended biases like those recently found in NLP evaluations and systems \citep[e.g.,][]{caliskan2017semantics,gururangan-18}.
Popular NLU benchmarks \citep[like SuperGLUE;][]{superglue} should be critically examined for potential \emph{construal biases}, and contrasts should be introduced deliberately to probe whether systems are modeling lexical choices, grammatical choices, and meaning in the desired way \cite{naik-18,kaushik-19,mccoy-19,gardner-20}.

%While the relevance of any particular dimension will likely be task-dependent, it will be instructive to know \emph{which} tasks benefit from \emph{which} dimensions. 

%For example, one could ask whether a proposed semantic representation captures the level of \textsc{resolution} at which entities and events are described, whether it represents \textsc{metaphor}, whether it is sensitive to the \textsc{prominence} of different participants or features of an event, and more.

As a broader suggestion, datasets should move away from a one-size-fits-all attitude based on gold annotations. Ideally, evaluation metrics should take into account not only partial structure matches, but also similarity to alternate construals.

\considercutting{
For example, one could ask the following questions of a proposed semantic representation: 

\begin{enumerate}
    \item Does it capture taxonomic relations for kinds of entities and events, as well as the level of \textsc{resolution} at which they are described?
    
    \item Does it represent \textsc{metaphor}, and if so, does it encode information about the source domain (e.g., \textsc{space}), the target domain (e.g., \textsc{time}), or both? 
    
    \item Is it sensitive to the \textsc{prominence} of different participants or features of an event, or is it intended to neutralize minor differences in prominence to focus on information content (\cref{sec:captioning})?
    
\end{enumerate}
}

%Recent research has revealed NLP evaluations and systems to exhibit a variety of unintended biases, including social biases and annotation artifacts \citep[e.g.,][]{caliskan2017semantics,gururangan-18}.
%A focus on certain aspects of meaning to the exclusion of others could be yet another form of bias.
%Popular benchmarks sometimes portrayed as testing the state of the art in NLU \citep[like SuperGLUE;][]{superglue} should be critically examined for \emph{construal biases}.
%; we anticipate questions such as the above may reveal new kinds of biases, or at least help clarify which aspects of meaning the evaluations are capable of testing.
%Questions such as the above can suggest contrasts to be introduced deliberately to probe whether systems are modeling lexical choices, grammatical choices, and meaning in the desired way \cite{naik-18,kaushik-19,mccoy-19,gardner-20}. 

\paragraph{Cognitive connections.} The many connections between construal and the rest of cognition highlight the need for further interdisciplinary engagements in the study of construal.
%are crucial to efforts to build systems with %human-level language capabilities.

%Closer collaboration with psycholinguistics could be mutually beneficial for both fields. 
% There is a large literature documenting the behavioral and neurological responses associated with construal, some of which we have reviewed here (see \cref{sec:dimensions}, \cref{sec:ops}). 

%could take advantage of the rich psycholinguistic literature on construal and directly model human data, using measures of human language processing or production as additional benchmarks for model evaluation.

The psycholinguistics literature is a particularly rich source of construal-related data and human language benchmarks. Psycholinguistic data could also be used to probe neural language models \citep{futrell2018rnns, linzen2018distinct, van2018modeling, ettinger2019bert}. How well do such models capture the phenomena reviewed in \cref{sec:dimensions}, and where do they fall short?

%In turn, these limitations serve as a test of psycholinguistic theory: which input channels or sources of world knowledge could augment those models so as to better match human data? 
%For example, could linguistically conveyed meaning be \emph{grounded} \cite{HARNAD1990335, kiros2018illustrative, bender-20, bisk-20} in existing or proposed embodied \cite{bergen2013embodied}, simulation-based \citep{tamari-20}, and/or frame-based \cite{torrent2018} representations? Also, would a system incorporating these representations produce results that better approximate human data? 

%Research pursuing implementations to answer those questions would contribute both to better understanding of the cognitive process involved in human linguistic capabilities, and to the development of more human-like linguistic agents. 

A fuller account of the constellation of factors involved in construal should also take seriously the grounded, situated nature of language use \cite{HARNAD1990335,kiros2018illustrative,bender-20,bisk-20}.  Frameworks motivated by the linguistic insights mentioned in \cref{sec:meaning} (such as the work on \textit{computational construction grammar} referenced in \cref{sec:ops}) and by growing evidence of \textit{embodied simulations} as the basis for meaning \cite{narayanan-00,bergen-chang-05,feldman-06,bergen-12,tamari-20} are especially relevant lines of inquiry.

Much work remains to flesh out the construal dimensions, operations and phenomena preliminarily identified in \cref{sec:dimensions} and \cref{sec:ops}, especially in connecting to typological, sociolinguistic, developmental, and neural constraints on conceptualization. 
We believe a concerted effort across the language sciences would provide valuable guidance for developing better NL systems and resources. %technologies, apps, resources

\section{Conclusion}

As the saying goes, the camera doesn't lie---but it may tell us only a version of the truth. The same goes for language. 

Some of the phenomena we have described may seem, at first glance, either too subtle to bother with or too daunting to tackle. But we believe it is both timely and necessary, as language technologies grow in scope and prominence, to seek a more robust treatment of meaning. We hope that a deeper appreciation of the role of construal in language use will spur progress toward systems that more closely approximate human linguistic intelligence.

\section*{Acknowledgments}

We are grateful to Lucia Donatelli, Nick Hay, Aurelie Herbelot, Jena Hwang, Jakob Prange, Susanne Riehemann, Hannah Rohde, Rachel Rudinger, and anonymous reviewers for many helpful suggestions; and to the ACL~2020 organizers for planning a special theme, \textit{Taking Stock of Where We’ve Been and Where We’re Going}. Special thanks to Nora Chang-Hay for finally relaxing her tiny iron grip.

This research was supported in part by NSF award IIS-1812778. The FrameNet Brasil Lab is funded by CAPES grants 88887.125411/2016-00 and 88887.144043/2017-00.

\bibliography{construal}
\bibliographystyle{acl_natbib}

% \appendix

\end{document}